# An RGB-D Image Dataset for Lychee Detection and Maturity Classification for Robotic Harvesting

Zhenpeng Zhang, Yi Wang, Shanglei Chai, Yingying Liu, Zekai Xie, Wenhao Huang, Pengyu Li, Zipei Luo, Dajiang Lu*, and Yibin Tian*

College of Mechatronics and Control Engineering, Shenzhen University, Shenzhen 518060, China;

zpzhang@szu.edu.cn (Z. Z.); 2410095042@mails.szu.edu.cn (Y. W.); 2450096004@mails.szu.edu.cn (S. C.);

liuyying@szu.edu.cn (Y. L.); 2410095049@mails.szu.edu.cn (Z. X.); huangwenhao2023@email.szu.edu.cn (W. H.);

2310294020@email.szu.edu.cn (P. L.); 2310295040@email.szu.edu.cn (Z. L.).

* Correspondence: ludajiang@szu.edu.cn (D. L.）; ybtian@szu.edu.cn (Y. T.)

**Abstract:** Lychee is a high-value subtropical fruit. The adoption of vision-based harvesting robots can significantly improve productivity while reduce reliance on labor. High-quality data are essential for developing such harvesting robots. However, there are currently no consistently and comprehensively annotated open-source lychee datasets featuring fruits in natural growing environments. To address this, we constructed a dataset to facilitate lychee detection and maturity classification. Color (RGB) images were acquired under diverse weather conditions, and at different times of the day, across multiple lychee varieties, such as Nuomici, Feizixiao, Heiye, and Huaizhi. The dataset encompasses three different ripeness stages and contains 11,414 images, consisting of 878 raw RGB images, 8,780 augmented RGB images, and 1,756 depth images. The images are annotated with 9,658 pairs of lables for lychee detection and maturity classification. To improve annotation consistency, three individuals independently labeled the data, and their results were then aggregated and verified by a fourth reviewer. Detailed statistical analyses were done to examine the dataset. Finally, we performed experiments using three representative deep learning models to evaluate the dataset. It is publicly available for academic use.

## 1. Background & Summary

Lychee is a high-value subtropical fruit, with China accounting for approximately 50% of global production [1]. Accurate detection of lychees is crucial for both yield estimation and harvesting during the fruit's ripening period [2-5]. As agricultural labor continues to decline, manual yield estimation and harvesting of lychees in complex terrains are becoming increasingly challenging. In this context, replacing traditional labor with intelligent technologies such as computer vision and robotics has emerged as a promising trend [6].

Some researchers have utilized vision algorithms for lychee detection and quality evaluation. For example, Xie et al. incorporated a small object detection layer into YOLOv8 and introduced a lightweight design using the mixed local channel attention module [7]. The optimized YOLOv8-BL network significantly improved the detection performance of inferior lychee fruits in orchard environments. Xiong et al. extracted the average hyperspectral data of the regions of interest of lychee fruits for spectral analysis [8]. Hyperspectral samples of newly- and slightly-damaged lychee fruits were selected, and a predictive model was established using partial least squares discriminant analysis, enabling quantification of different levels of fruit quality. Other researchers have employed visual methods for lychee yield estimation. Peng et al. introduced an additional small object detection layer to enhance the feature extraction of lychees in UAV imagery and integrated the efficient channel attention mechanism [9]. They proposed an improved model named YOLO-Lychees, which demonstrated enhanced robustness and higher detection accuracy. This improved model is well-suited for lychee identification and yield estimation in field environments. Wang et al. improved the YOLOv5s backbone by integrating ShuffleNet-v2, and introduced the CBAM module during the feature fusion stage [10]. Their results demonstrated that the model performed well in terms of accuracy and robustness across diverse conditions. Additionally, some researchers have utilized visual methods for lychee maturity classification, which can further enable collaboration with intelligent robots to achieve automated harvesting. Liang et al. integrated a hybrid convolutional-transformer search module into YOLOv8 to perform instance segmentation of lychees in images, obtaining both fruit objects and masks [11]. They applied the KGAP-DBSCAN clustering algorithm, where KGAP refers to K-distance gap and DBSCAN stands for Density-Based Spatial Clustering of Applications with Noise, to determine the clustering radius based on the density of object points, enabling automatic clustering of fruits into bunches. The proposed method achieved strong performance in maturity grading of individual fruits, with a precision of 94.20% and a recall of 91.91%. Liang et al. proposed a method based on UAV remote sensing and YOLOv8-FPDW [12], which integrates FasterNet, ParNetAttention, DADet, and Wiou modules for fast and accurate classification of lychee maturity. The improved model demonstrated robust performance under various field conditions. Integrating multimodal perception with specialized robotic hardware for lychee harvesting has also been investigated. Yao et al. designed a vision–tactile fusion-based lychee comb-cutting end-effector, considering the characteristics of lychee trees such as branching and clustered fruit growth [13]. A zero-shot prediction large model combined with the Hough circle detection method was used for lychee detection in the image, while tactile sensing between the gripper fingers was used to update the grasping state of the fruit stems in real time. The integrated system enabled the harvesting robot to perceive the fruit stems. Experimental results showed that the static tactile stem perception rate was only 16.7%, whereas the vision–tactile fusion approach increased this rate to 86.7%.

Deep learning (DL) has demonstrated performance in accuracy and speed comparable to those of human experts in several fruit detection and harvesting tasks [14]. However, most existing studies rely on private datasets [15, 16], which are often limited by small sample sizes, inconsistent fruit labeling and maturity grading standards, and significant variations in image acquisition protocols. A core challenge in this field remains the lack of standardized, publicly available datasets. We conducted an extensive search for open-source lychee datasets; however, none of them meets the criteria for a standardized benchmark. For instance, some datasets released by researchers consist only of single-lychee images [17-19], which offer limited practical utility for lychee detection and harvesting tasks. The VNU dataset includes only two categories: ripe and unripe lychees, and also confounds the annotations with lychee flower labels [20] . MeoMeo complied a dataset by collecting lychee images from the internet [21], but it suffers from noise, annotation errors, and inconsistent scene settings. Zhao provided an on-tree lychee dataset with four maturity levels (immature, green, semi-ripe, and fully ripe) [22] , but more than 500 images depict single or sparsely distributed fruits, and over 800 images exhibit severe distortions. Moreover, none of these datasets include challenging environmental conditions such as evening or rainy scenarios. They only provide color images with object detection annotations, and lack depth information. These limitations severely hinder the objective comparison of algorithm performance and impede further technological innovation for harvesting robots.

To address these challenges, we constructed a new lychee dataset featuring more than one data modality, large intra-class variations, consistent annotation criteria, and well-defined maturity categories. Compared with existing datasets, as shown in Table 1, our dataset introduces the following improvements: (1) It comprises 11,414 high-resolution images of multiple lychee varieties at different maturity stages, collected from the major lychee-producing region in China. This ensures broader varietal representation and diverse weather and environmental conditions; (2) It provides high-quality, consistent annotations for all images; (3) The dataset offers enhanced diversity through temporal and scene-based splits, enabling flexible combination for different application scenarios; (4) In addition to raw color images, it also includes augmented images and corresponding depth maps.

The dataset is released in an open-source format. It aims to facilitate standardized research in computer vision algorithms for lychee detection and maturity classification, and to provide robust data support for the real-world deployment of lychee harvesting robots.

Table 1. Comparison of open-source lychee datasets. (T: train; V: validation; C: color; D: depth; Rect: rectangle)

| Dataset | Condition | Environment | Image | Class | Resolution | Annotation |
|---|---|---|---|---|---|---|
| HUST [17] | Single withered fruit | Indoor on-table | 512 (C) | 1 | 640×640 | Rect |
| SUC [18] | Single plastic fruit | Indoor on-table | 228 (C) | 2 | 640×640 | Polygon |
| Mihai [19] | Single fruit | Blank background | 490 (C) | 0 | 100×100 | Rect |
| Vnu [20] | Mostly immature fruit | Outdoor on-tree | 1005 (C) | 3 | 640×360 | Rect |
| MeoMeo [21] | Peeled and unpeeled fruit | From the Internet | 956 (C) | 2 | 640×640 | Rect |
| Zhao [22] | Four stages of ripeness | Outdoor on-tree | 1308 (C, T)<br>341 (C, V) | 4 | 640×640 | Rect + Maturity |
| Ours | Weather: Sunny, rainy<br>Time: morning, noon, evening<br>Period: over 3 weeks | Indoor on-table<br>Outdoor on-tree | 9658 (C)<br>1756 (D) | 3 | 1280×1024 | Rect + Maturity |

## 2. Methods

The main steps of dataset preparation are illustrated in Fig. 1. First, raw images were collected using a customized sensor module. Next, depth maps were obtained from the RGB images by DL-based monocular depth estimation. Then, to enhance data diversity, various data augmentation operations were carried out to increase the sample size. Afterwards, the images underwent qualitative and quantitative analysis from multiple dimensions. Subsequently, the images were annotated for single lychee detection and maturity classification. Finally, three representative DL models were employed to comprehensively verify the dataset.

2.1. Image Acquisition

2.1.1 Sensor Module

A multimodal sensor module was built in-house, as shown in Fig. 2(a). It mainly consists of a color (RGB) camera with 1280×1024 pixels (MV-CU013-A0GC by Hikvision, Hangzhou, China), a thermal (IR) camera with 640×512 pixels (MV-CI003-GL-N6 by Hikvision, Hangzhou, China), and a Lidar (Mid70 by Livox, Shenzhen, China), and an edge-computing SoC (Jetson NX by Nvidia, Santa Clara, USA). It was originally designed to be mounted on a drone to carry out lychee data collection, and the Lidar was intended to be used for drone guidance and obstacle avoidance. However, due to some regulatory challenges, drones were not used for data collection in this study, and the sensor module was mounted on a metal pole that was carried by an operator for image acquisition, as illustrated in Fig. 2(b). The sensor module was operated by the Robot Operating System (ROS) [23]. The cameras were automatically calibrated with a checkerboard target using a procedure similar to that in previous studies [24].

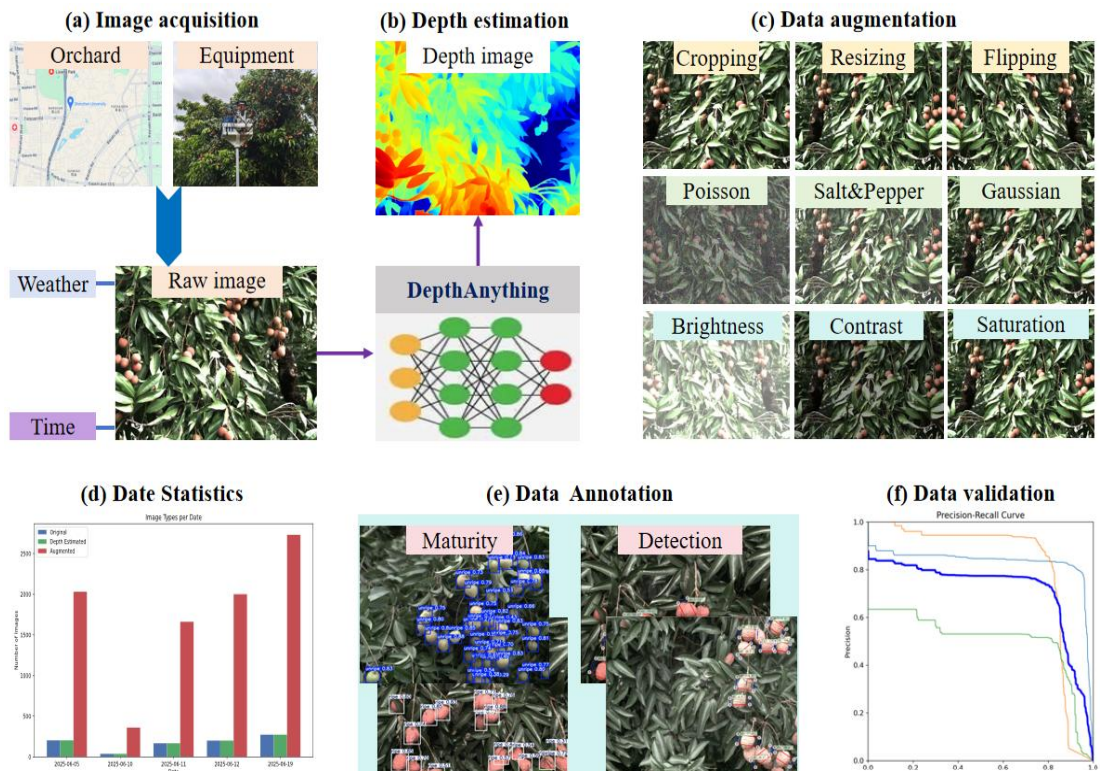

**Fig. 1.** The main steps of dataset preparation. (a) Image acquisition at different orchard locations, under different weather conditions and at different times of the day. (b) Monocular depth estimation from the RGB images using a deep neural network. (c) Data augmentation by geometrical transformations, adding noise and pixel value adjustments. (d) Obtaining data records and statistics. (e) Data annotation for lychee fruit detection and maturity classification. (f) Data validation using multiple existing neural networks.

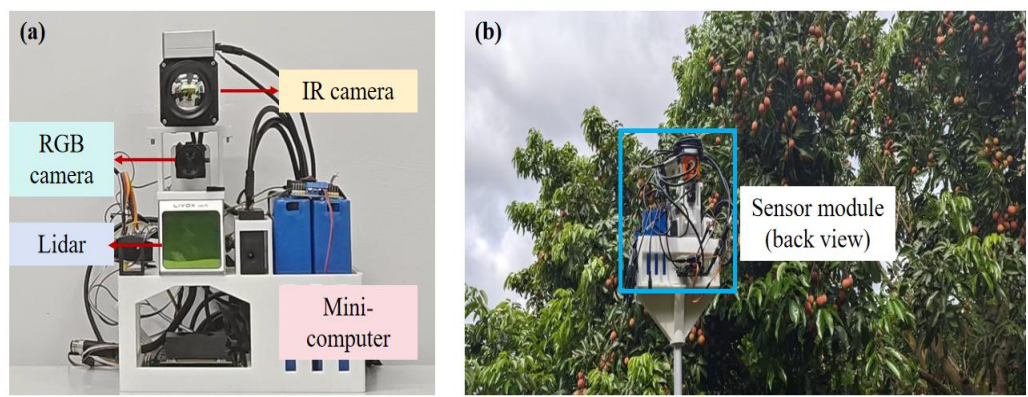

**Fig. 2.** The multimodal sensor module for image acquisition. (a) The layout of the main components. (b) The sensor module mounted on the top of a pole during image acquisition in front of a lychee tree.

As the thermal images showed significant inconsistency under different environmental conditions (as illustrated in Fig. 3), they were not included in this study. We will further address the thermal imaging consistency issue in future work. For the current dataset, only images from the RGB camera are utilized. Its detailed parameters are shown in Table 2.

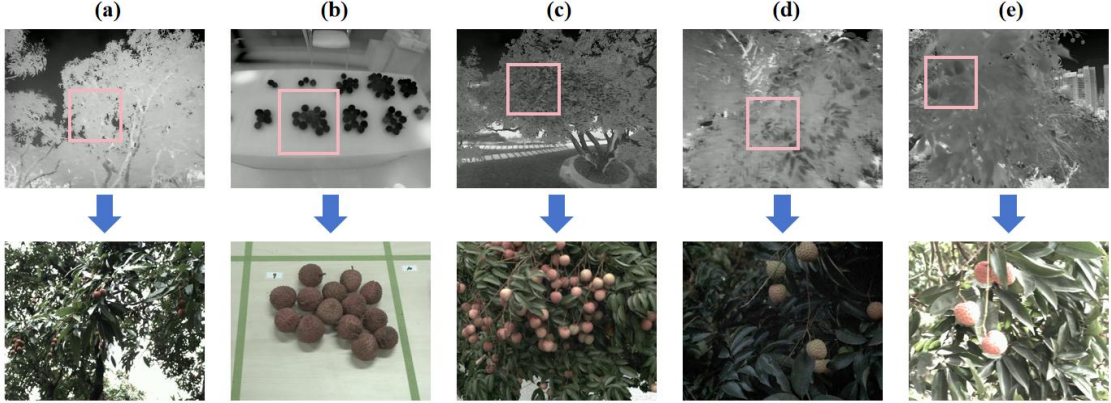

**Fig. 3.** (a-e) Thermal infrared images under five different environmental conditions exhibit noticeable inconsistencies. Top: Infrared images, where the pink rectangles indicate lychee regions. Bottom: Corresponding RGB images of the corresponding rectangular areas.

**Table 2.** Parameters of the RGB camera.

| Parameter | Value | Note |
|---|---|---|
| Model | MV-CU013-A0GC | Industrial camera |
| Resolution | 1280×1024 | |
| Pixel size | 4.8 μm | |
| Max frame rate | 91.3 fps @ 1280×1024 | Supports high-speed image acquisition |
| Dynamic range | 54 dB | Supports a variety of light conditions |
| Exposure time | 10 μs to 10 sec | Supports auto exposure; Support short exposure to reduce motion blur |
| Focal length (label) | 12mm | |
| Field of view | ~ $29^0 \times 23^0$ | Computed from focal lengths and sensor size |
| Focal length (pixel) | 2527.7411, 2527.4279 | Horizontal and vertical focal lengths $f_x$, $f_y$ |
| Optical center (pixel) | 636.5080, 484.1510 | Horizontal and vertical center $c_x$, $c_y$ |
| Radial distortion | -0.0223, 0.7687, -4.1053 | $k_1$, $k_2$, $k_3$, usually caused by lens shape |
| Tangential distortion | -0.0008, -0.0005 | $p_1$, $p_2$, usually caused by lens and sensor misalignment |

2.1.2. Image Acquisition Procedure

Videos were captured from one side of lychee trees at a distance of roughly 20–60 cm from the canopies, with a walking speed of approximately 1 m/s. The low-speed movement helped reduce motion-induced image blur. The orchards were located at two nearby sites in Shenzhen, as shown in Fig. 4, specifically at the Yuehai campus Shenzhen University and in the Xili subdistrict. Data collections were conducted during the three-week peak of lychee ripening (specifically, on June 5, 10, 11, 12, and 19, 2025). Multiple common lychee varieties, such as Nuomici, Feizixiao, Heiye, and Huaizhi, were captured under various weather conditions (e.g., light rain on June 11, and sunny on other days) and at different times of day (morning, noon, and evening).

The ROS plugin (bag_to_images) was used to convert the recorded .bag files into image sequences. One image was extracted for every 10 frames in the video. To improve dataset diversity and reduce redundancy, after the first image was manually selected in a sequence, the Structural Similarity Index (SSIM) [25] was employed to select the most dissimilar frames within each 10-frame segment from the previous selected image.

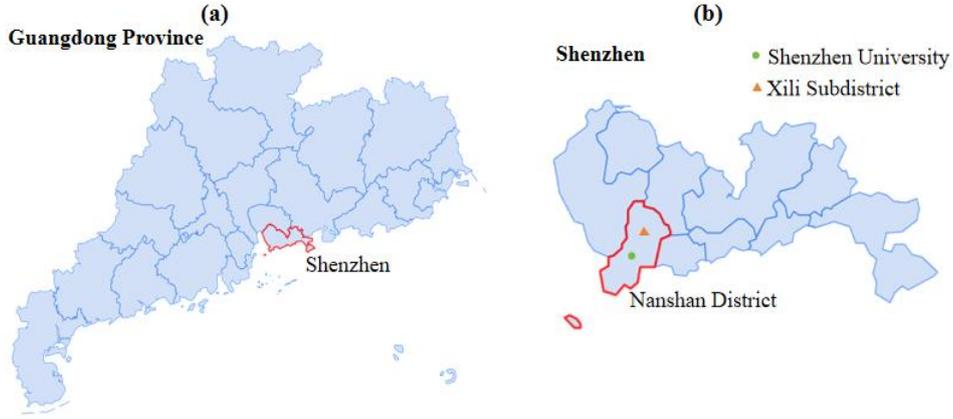

**Fig. 4.** The locations of lychee image acquisition. (a) The lychee orchard on the Yuehai campus of Shenzhen University, where the cultivars include Nuomici, Feizixiao, Huaizhi, and Baila; (b) The Xili Subdistrict lychee orchard, where the cultivars include Guiwei, Heiye, and Zazhi.

2.2. Monocular Depth Estimation

For robotic harvesting, depth information is important for grasping. In addition, a number of previous studies on cherry tomato detection have shown that incorporating depth also improves fruit detection performance [26-28]. We employed the DepthAnything deep neural network to estimate relative depth from the raw color images [29], as illustrated in Fig. 1(a-b). This model was trained jointly on both annotated and unlabeled data, resulting in stronger generalization capabilities, faster inference speed, fewer parameters, and higher depth estimation accuracy.

2.3. Data Augmentation

The generalization ability of a dataset greatly impacts the performance of trained DL models. Data augmentation is utilized to generate more diverse image samples, thereby minimizing the gap between the training set, validation set, and any test sets due to sampling [30].

2.3.1. Geometric Transformations

Scaling, cropping, horizontal and vertical flipping, and rotation were utilized geometric transformations. These transformations usually preserve annotations, as they only change the position of key features [31].

2.3.2. Addition of Noise

Noise can impair the performance of deep neural networks and lead to poor generalization. Their robustness can be improved by augmenting the data with different types of noise [32]. Gaussian noise, salt and pepper noise, and Poisson noise were added to the raw images to augment the dataset.

2.3.3. Pixel Value Transformation

Image pixel values were adjusted through matrix operations to randomly increase or decrease image brightness, contrast, and saturation [33].

# 3. Data Records

3.1. Data Statistics

This subsection reports the numbers of raw RGB images and filtered images collected on five different dates, followed by a detailed statistical analysis of the SSIM and Kernel Density Estimation (KDE) [34] results of different image groups. We calculated the SSIMs for the images in each date-specific folder, and applied KDE using a Gaussian kernel to estimate the distribution of SSIM scores. Specifically, the mean and variance of SSIMs, as well as the CIs (confidence intervals) of KDE, were computed, with detailed results provided in Table 3.

**Table 3.** Statistics for images collected on different days.

| Date | Weather | Raw image | Selected image | SSIM mean | SSIM variance | KDE 95% CI | Note |
|---|---|---|---|---|---|---|---|
| June 5 | Sunny | 1107 | 203 | 0.0290 | 0.0004 | [0.0032, 0.0769] | Low SSIM. |
| June 10 | Indoor | 238 | 36 | 0.2136 | 0.0057 | [0.1091, 0.3976] | Moderate SSIM, broad distribution. |
| June 11 | Rainy | 1012 | 166 | 0.0862 | 0.0079 | [0.0479, 0.2184] | Mixed low and high SSIM. |
| June 12 | Sunny | 1190 | 200 | 0.1068 | 0.0029 | [0.0381, 0.2476] | Overall low SSIM, slightly broad. |
| June 19 | Sunny | 1520 | 273 | 0.0401 | 0.0005 | [0.0088, 0.1007] | Overall low SSIM, concentrated. |
| **Total/Ave** | - | **5067** | **878** | **0.0951** | **0.00348** | **[0.0031, 0.3976]** | **Overall low SSIM.** |

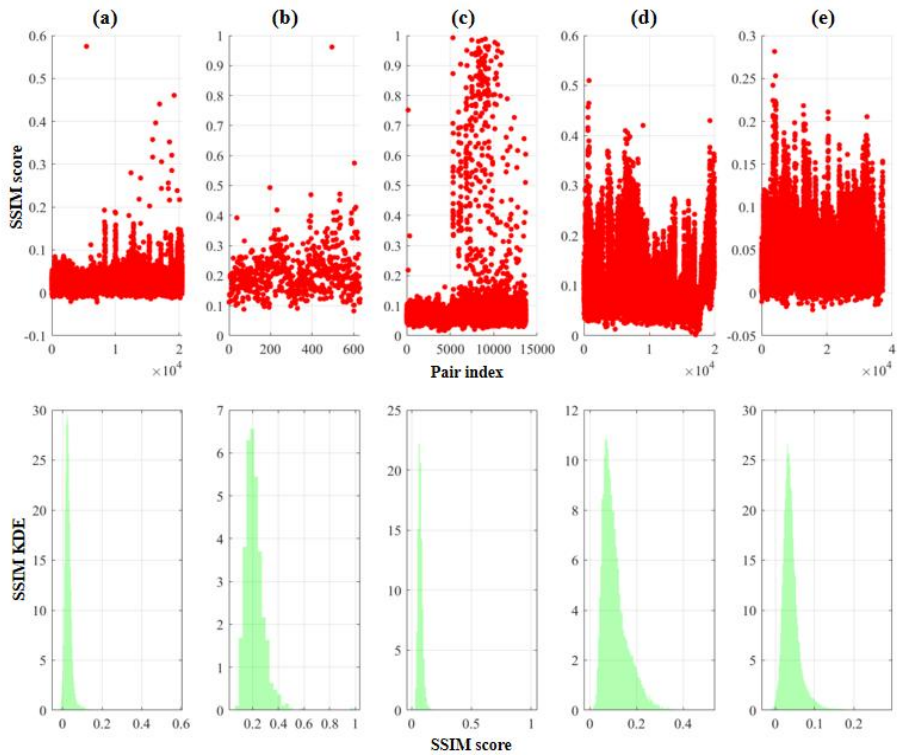

**Fig. 5.** Similarity (SSIM) analysis of selected lychee images. (a) June 5, 2025; (b) June 10, 2025; (c) June 11, 2025; (d) June 12, 2025; (e) June 19, 2025.

The SSIMs from the selected images on five different dates are shown using scatter plots (Fig. 5 top row) and KDE plots (Fig. 5, bottom row). The mean SSIM values across the five groups ranged from 0.03 to 0.21, all below 0.3, indicating generally low similarity among image pairs. The June 11 group exhibited the highest variance (0.0079), suggesting the presence of image pairs with substantial differences. In contrast, the June 5 group, which had the lowest mean SSIM, showed a variance of only 0.0004, implying that nearly all image pairs fell within the low-similarity range. The upper bound of the 95% KDE CI was approximately 0.21 for most groups, indicating that highly similar image pairs were rare and predominantly concentrated in the 0.0-0.2 range. Notably, the June 10 group had the highest mean SSIM, with its KDE distribution extending to an upper bound of 0.3976, suggesting the presence of a subset of moderately to highly similar image pairs.

Fig. 6 shows the statistics of the raw, augmented, and depth images in the final dataset. Fig. 6(a) presents the proportion of all images across the five dates, including raw and augmented RGB images, and depth maps. Fig. 6(b) shows the numbers of raw and augmented RGB images, and depth maps across the five dates. The analysis revealed an uneven data distribution. Images from June 10, 2025, were the fewest, accounting for only 4.1%, while images from June 19, 2025, were the most abundant, representing 31%. Augmented images comprised the majority of the dataset (approximately 77%). The dataset augmentation strategy resulted in a balanced distribution. During training, the proportion of augmented images can be considered, or the use of all augmented data can be adjusted according to the specific task requirements.

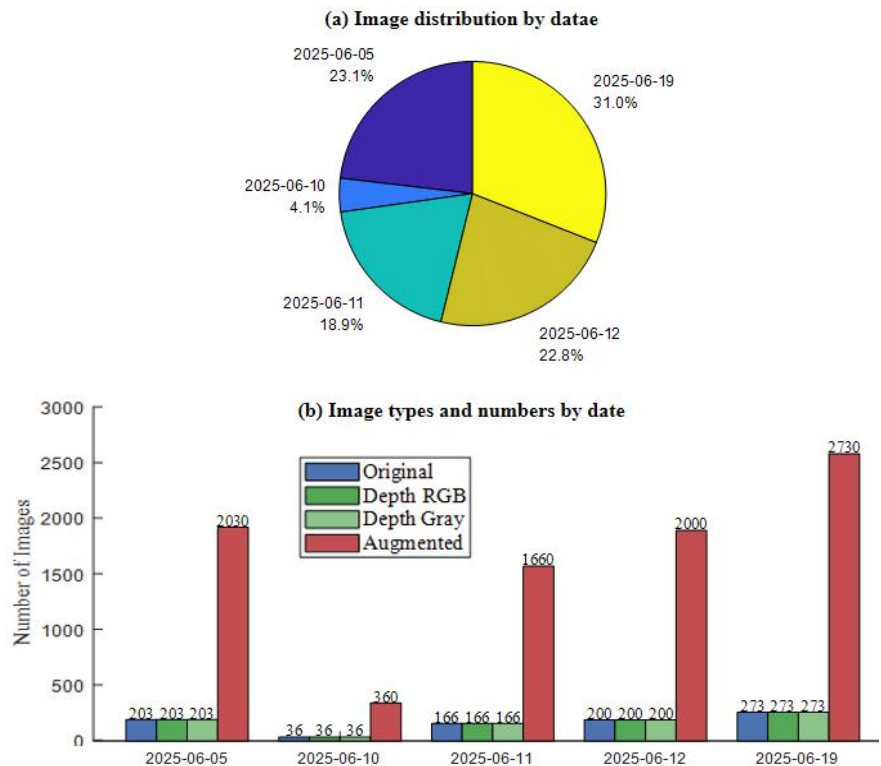

**Fig. 6.** Statistics of the dataset. (a) Image percentage by date; (b) Image types and numbers by date.

3.2. Data Annotation

The dataset is intended to advance lychee detection and harvesting in precision agriculture. Four of the authors spent about two months annotating the images for lychee detection (rectangular bounding boxes) and maturity classification. Before the annotation, each member received proper training in annotation techniques. The annotations were initially performed by three individuals, and any inconsistencies were resolved by the fourth senior member to ensure consistency. The overall annotation process is illustrated in Fig. 7.

Direct annotation of picking posture is still rare in existing fruit datasets [35]. We only annotated vertical rectangles for lychee fruits for this report. For maturity classification, annotations were conducted using LabelImg (v1.8.6) [36] and X-AnyLabel (v3.0.3) [37], with all samples categorized into three maturity levels: unripe, semi-ripe, and ripe.

For future work, the fruit pose can also be annotated, for example, using rotated rectangles instead of vertical ones, the most commonly used formats being Cornell [38] and Jacquard [39]. Unfortunately, the grasping datasets created by Häni et al. [40] and James et al. [41] did not release their annotation tools. There is a MATLAB-based tool that supports the Jacquard format [42], but it proved to have poor usability in practice. We plan to develop a new annotation tool that supports both Cornell and Jacquard formats, enabling more efficient and accurate labeling for picking posture.

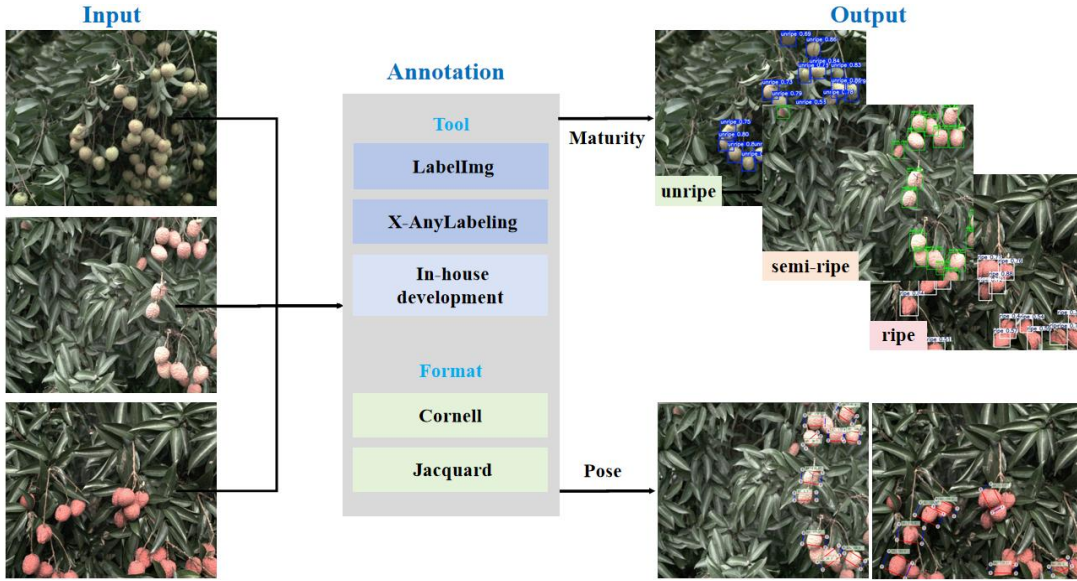

**Fig. 7.** Lychee image annotations for maturity classification and grasping pose detection.

Fig. 8 shows the number and area of maturity annotations. Fig. 8(a) illustrates the distribution of the annotation counts for the three maturity categories and their total. Among them, June 11 had the largest number of annotations, whereas June 10 had the fewest. For the unripe category, June 12 contained the most annotations, while June 10 contained the least. For the semi-ripe and ripe categories, June 11 had the highest number of annotations, and June 10 had the lowest. Fig. 8(b) presents the average and total pixel areas (MA and TA respectively) of annotations for each category and for all categories combined, calculated according to Eqs.1-2. The overall average annotation area was the largest on June 10 and the smallest on June 11. For the ripe category, the largest average annotation area occurred on June 10, and the smallest on June 5. To further perform a local analysis of annotation counts, Fig. 9 shows scatter plots of the three-category annotations per image on each date. The x-axis denotes the dates, the y-axis the image indices, and the z-axis the annotation counts. The maximum, minimum, and mean annotation counts per category per date were computed, with the values displayed on the z-axis representing the means. The analysis indicates that immature annotations were relatively dispersed across all dates, whereas ripe annotations were more concentrated. Semi-ripe and ripe annotations were particularly abundant and concentrated on June 11.

$$MA_c = \frac{\sum_i^{N_c}(W_i \cdot H_i)}{N_c} \tag{1}$$

$$TMA = \frac{\sum_i^{N_c} MA_c}{\sum_{c=1}^{3} N_c} \tag{2}$$

where $N_c$ denotes the number of annotation boxes of the category on each date, $W_i$ and $H_i$ are the normalized width and height of the annotation boxes, respectively.

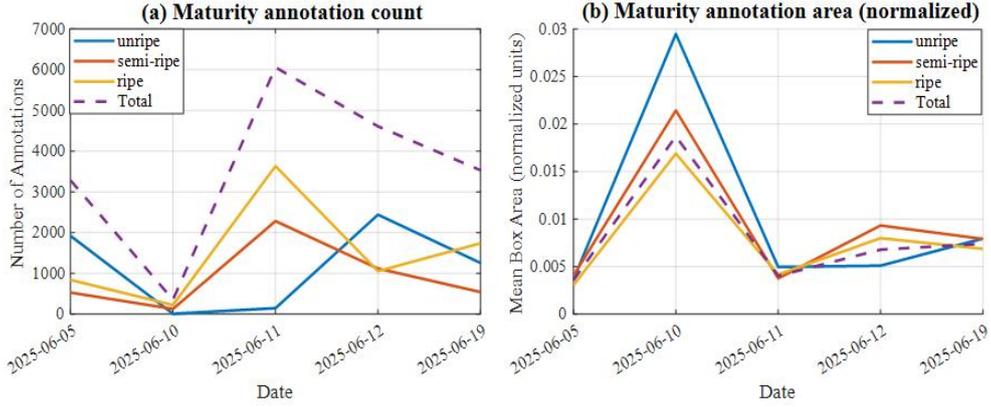

**Fig. 8.** Maturity annotation statistics across five dates. (a) Annotation counts for the three maturity categories; (b) Average annotation areas (normalized).

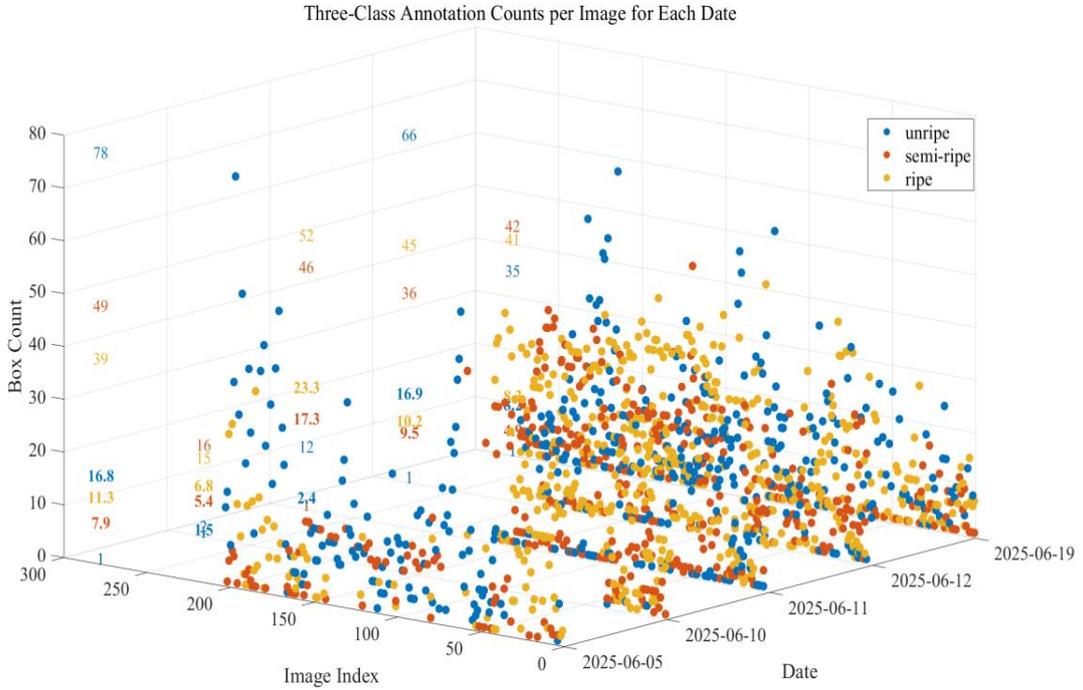

**Fig. 9.** Three maturity level annotation counts per image for each date.

3.3. Dataset Splitting

As shown in Table 4, we mixed the raw and augmented images of maturity detection from the same date, randomly shuffled the image order, and split them into training, validation, and test sets with a ratio of 8:1:1. This design allows users to flexibly select subsets according to their task requirements and computational resources. For example, on devices with limited computational capacity, subsets from individual dates can be used independently, whereas on devices with sufficient resources, multiple subsets or the entire dataset can be combined for training.

**Table 4.** The dataset distribution for maturity detection. (Aug: augmentation)

| Date | Train (Raw/Aug) | Validate (Raw/Aug) | Test (Raw/Aug) |
| --- | --- | --- | --- |
| June 5 | 162/1624 | 20/203 | 21/203 |
| June 10 | 28/288 | 4/36 | 4/36 |
| June 11 | 132/1328 | 16/166 | 18/166 |
| June 12 | 160/1600 | 20/200 | 20/200 |
| June 19 | 218/2184 | 27/273 | 28/273 |
| **Total** | **700/7024/** | **87/878** | **91/878** |

# 4. Technical Validation

4.1. Experimental Setup

All experiments were conducted on a workstation equipped with an Intel i7 CPU and an NVIDIA GeForce RTX 4090 GPU (Santa Clara, USA). The DL algorithms were implemented in CUDA 12.1, Python 3.8.10, and Pytorch 2.4.1. During training, all images were resized to 640×640. The batch size was set to 16, and the number of training epochs was 1000, patience was set to 100 (epochs to wait for no observable improvement for early stopping of training). The initial learning rate was set to 0.01 and decayed to 1% of the initial value. Stochastic Gradient Descent (SGD) [43] was used as the optimizer, with a momentum of 0.937. A patience value of 100 and a weight decay factor of 0.0005 were applied to regularize the model parameters. To enhance convergence and performance, a cosine annealing learning rate scheduler was used to dynamically adjust the learning rate throughout training.

4.2. Evaluation Metrics

All experimental evaluations were conducted on the test set, analyzing both the class-wise performance and the overall average performance across all categories. The evaluation metrics used follow standard practices commonly adopted in other fruit detection studies [27]. Precision measures the proportion of correctly predicted positive samples (i.e., correctly detected targets) among all predicted positives. Recall represents the proportion of true positive samples that are correctly detected, reflecting the model's ability to minimize missed detections. For each class, we report the average precision (AP) as the primary evaluation metric. Additionally, the mean AP (mAP) across all classes is reported. AP for each class is computed over the standard IoU thresholds from 0.5 to 0.95 with a step of 0.05, with AP specifically reported at IoU = 0.5 and across 0.5–0.95. We also report the number of model parameters and memory usage as supplementary performance indicators. The F1 score, the harmonic mean of precision and recall, is also reported to balance the trade-off between these two metrics [44].

4.3. Lychee Detection and Maturity Classification Results

We conducted maturity detection experiments using the data partitions of June 5, June 11, June 12, and June 19 from Table 4. Two sets of experiments were performed: one using only the raw images and the other using both the raw and augmented data. The DL models included RT-DETR-ResNet50 [45], hereafter referred to as RT-D-Res, and the more recent YOLO models (YOLOv8n and YOLOv12n) [46] for comparison. Fig. 11 presents the comparison results on the same validation set derived from the raw data. With data augmentation, the precision-recall (PR) curves shifted toward the upper-right corner, indicating improved performance across all models. Table 5 compares the evaluation metrics before and after augmentation, where the underlined values highlight the improvements achieved with augmented data. On average, models of the same architecture trained with augmentation achieved approximately 10%-20% higher mAP50 and mAP compared with those trained without augmentation.

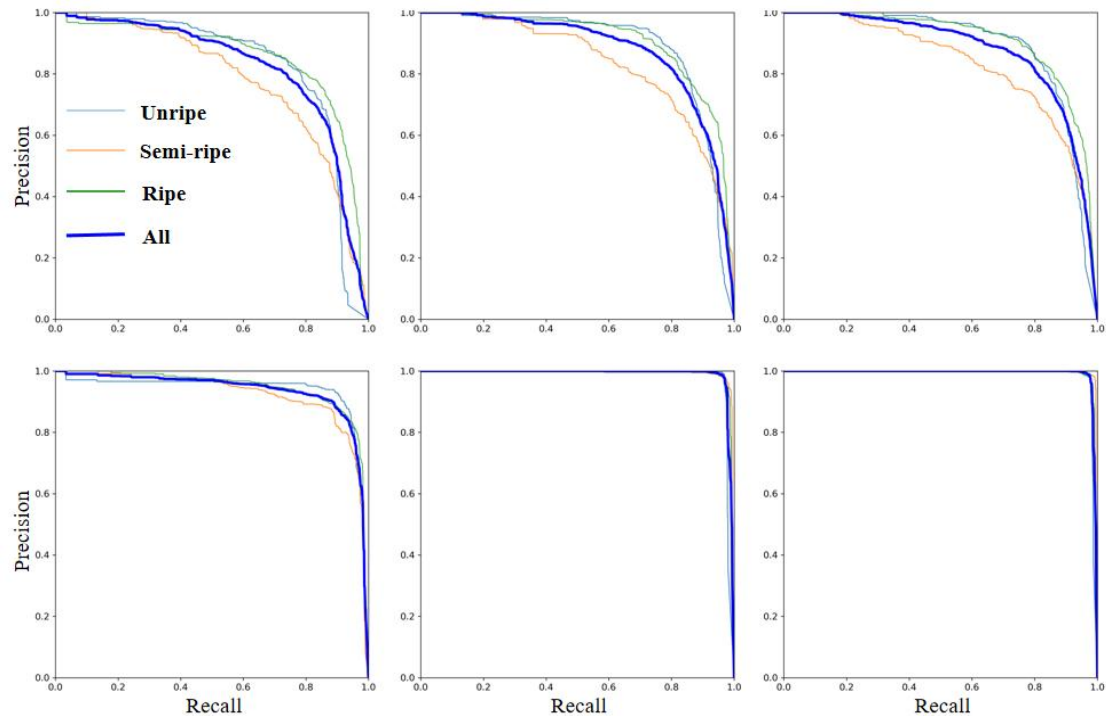

**Fig. 11.** (a)-(c) Precision-Recall curves of RT-D-Res, YOLOv8n, and YOLOv12n on the test data. The top and bottom rows are for models trained without and with augmented data, respectively.

Fig. 12 shows the confusion matrices of the three DL models for the lychee maturity classification task, evaluated without and with data augmentation. The comparison highlights that data augmentation leads to a noticeable improvement in classification performance across all models, particularly in reducing the misclassification between semi-ripe and ripe samples. Without augmentation, a substantial proportion of semi-ripe samples were incorrectly predicted as ripe, resulting in relatively low accuracies of 68.44%, 72.49%, and 68.25% for RT-D-Res, YOLOv8n, and YOLOv12n, respectively. In contrast, the unripe class maintained a consistently high recognition rate (around 94%) across all models, indicating its more distinctive features. After data augmentation, the classification accuracies improved markedly, with RT-D-Res and YOLOv8n showing stable performance across all three classes. Their semi-ripe accuracies increased to 88.86% and 88.82%, while the ripe class reached 93.82% and 94.38%. Although YOLOv12n also benefited from augmentation, its semi-ripe accuracy (81.14%) remained lower than that of the other two models, despite achieving a reasonable result (85.12%) on the ripe class. These results confirm that the dataset augmentation strategy is effective in improving the robustness and discriminative ability of different classification models.

Table 5. Evaluation metrics of lychee maturity detection models before and after data augmentation

| Category | DNN | Image | Metrics | | | | |
| --- | --- | --- | --- | --- | --- | --- | --- |
| | | | Precision | Recall | mAP50 | mAP50-95 | GFLOPs |
| All | RT-D-Res | Raw | 0.753 | 0.783 | 0.821 | 0.674 | 125.6 |
| | | Raw+Aug | 0.878 | 0.907 | 0.936 | 0.768 | 125.6 |
| | Yolov8n | Raw | 0.800 | 0.797 | 0.874 | 0.733 | 8.1 |
| | | Raw+Aug | 0.981 | 0.965 | 0.989 | 0.941 | 8.1 |
| | Yolov12n | Raw | 0.815 | 0.777 | 0.875 | 0.730 | 6.3 |
| | | Raw+Aug | 0.983 | 0.979 | 0.992 | 0.951 | 6.3 |
| Unripe | RT-D-Res | Raw | 0.755 | 0.802 | 0.828 | 0.681 | 125.6 |
| | | Raw+Aug | 0.919 | 0.905 | 0.939 | 0.759 | 125.6 |
| | Yolov8n | Raw | 0.785 | 0.846 | 0.889 | 0.743 | 8.1 |
| | | Raw+Aug | 0.981 | 0.970 | 0.982 | 0.939 | 8.1 |
| | Yolov12n | Raw | 0.814 | 0.846 | 0.888 | 0.738 | 6.3 |
| | | Raw+Aug | 0.992 | 0.972 | 0.987 | 0.950 | 6.3 |
| Semi-ripe | RT-D-Res | Raw | 0.731 | 0.707 | 0.780 | 0.632 | 125.6 |
| | | Raw+Aug | 0.850 | 0.891 | 0.924 | 0.764 | 125.6 |
| | Yolov8n | Raw | 0.793 | 0.706 | 0.829 | 0.690 | 8.1 |
| | | Raw+Aug | 0.978 | 0.965 | 0.992 | 0.940 | 8.1 |
| | Yolov12n | Raw | 0.805 | 0.663 | 0.833 | 0.692 | 6.3 |
| | | Raw+Aug | 0.978 | 0.991 | 0.995 | 0.953 | 6.30 |
| Ripe | RT-D-Res | Raw | 0.773 | 0.839 | 0.854 | 0.708 | 125.6 |
| | | Raw+Aug | 0.863 | 0.923 | 0.946 | 0.783 | 125.6 |
| | Yolov8n | Raw | 0.822 | 0.839 | 0.906 | 0.767 | 8.1 |
| | | Raw+Aug | 0.984 | 0.960 | 0.992 | 0.942 | 8.1 |
| | Yolov12n | Raw | 0.826 | 0.822 | 0.904 | 0.761 | 6.3 |
| | | Raw+Aug | 0.979 | 0.973 | 0.993 | 0.950 | 6.3 |

Fig. 13 visually illustrate the detection and classification results of the three models on a randomly selected sample from the dataset. Red circles indicate lychees of misclassified maturity. It reveals that in sparse regions, the localization ability of the models showed little difference before and after augmentation. However, in dense regions with small lychees, the non-augmented models produced misclassifications with lower recognition and accuracy, whereas the augmented models avoided these issues. Overall, the results demonstrate that the data augmentation method effectively improves the comprehensive performance of the models.

# 5. Data Availability

The dataset is available at：https://github.com/SeiriosLab/Lychee．The Python scripts for data augmentation, image similarity comparison, and annotation are available within the same repository under the tree/main/script directory.

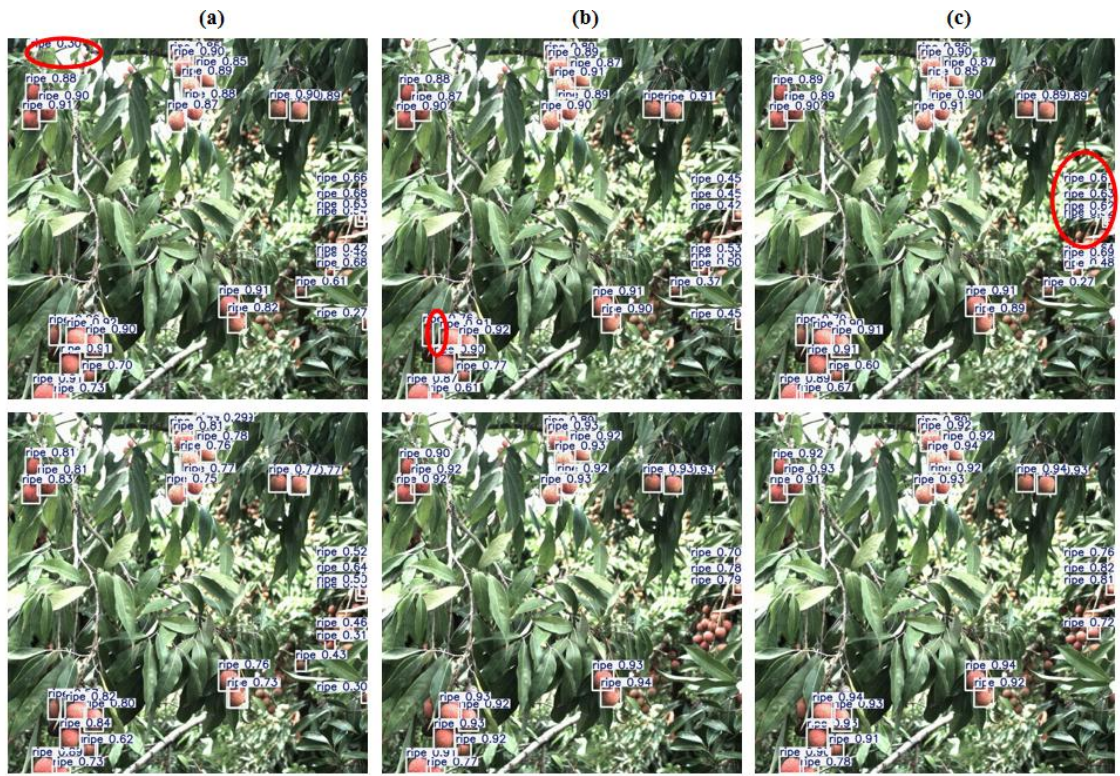

**Fig. 12.** Confusion matrices of maturity classifications. (a-c) The results of RT-D-Res, YOLOv8n, and YOLOv12n, respectively. The top and bottom rows show the results without and with data augmentation, respectively.

**Fig. 13.** Lychee detection and maturity classification results on a randomly selected sample. (a-c) The results of RT-D-Res, YOLOv8n, and YOLOv12n, respectively.

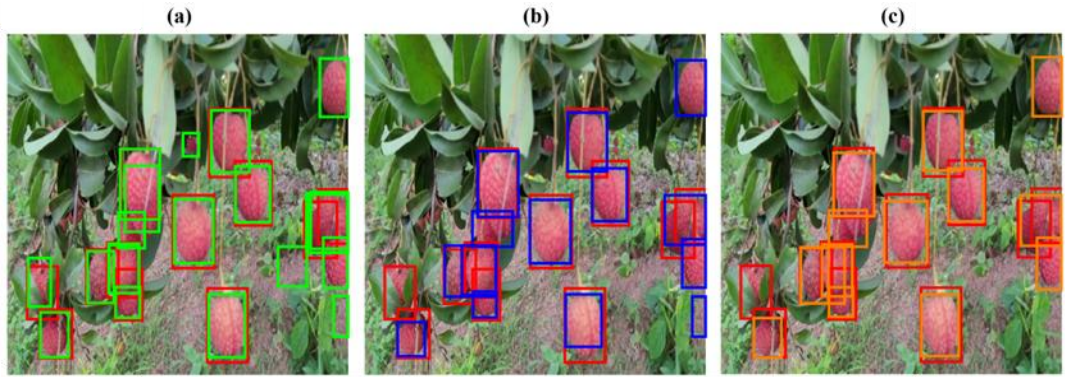

**Fig. 14.** Detection results of the three DL models on an example from an external dataset. (a-c) The detection results of RT-D-Res, YOLOv8n, and YOLOv12n, trained with augmented data, respectively.

# Acknowledgement

The study was partially funded by Shenzhen University (No. 20230300).

# Author Contributions

Zhenpeng Zhang, Yi Wang and Shanglei Chai: research design, investigation, data collection, data labeling, and drafting the manuscript; Yingying Liu, Zekai Xie: data curation, software, data labeling, validation; Wenhao Huang, Pengyu Li, Zipei Luo: equipment, data collection; Dajiang Lu: data validation and verification; Yibin Tian: conceptualization, visualization, manuscript revision. All author agreed with final manuscript.

# Competing Interests

The authors declare no competing interests.